\pdfoutput=1

\documentclass[11pt]{article}

\usepackage{acl}

\usepackage{times}
\usepackage{latexsym}
\usepackage{graphicx}
\usepackage{amsmath}
\usepackage{dblfloatfix}

\usepackage[T1]{fontenc}

\usepackage[utf8]{inputenc}

\usepackage{microtype}

\title{Partially Randomizing Transformer Weights for Dialogue Response Diversity}

\author{Jing Yang Lee\textsuperscript{1}, Kong Aik Lee\textsuperscript{2}, Woon Seng Gan\textsuperscript{3}\\\\
  School of Electrical and Electronic Engineering, Nanyang Technological University\textsuperscript{1,3} \\
  Institute for Infocomm Research, A*STAR\textsuperscript{2}\\
  jingyang001@e.ntu.edu.sg\textsuperscript{1}, lee\_kong\_aik@i2r.a-star.edu.sg\textsuperscript{2}, ewsgan@ntu.edu.sg\textsuperscript{3} \\}

\begin{document}
\maketitle
\begin{abstract}
Despite recent progress in generative open-domain dialogue, the issue of low response diversity persists. Prior works have addressed this issue via either novel objective functions, alternative learning approaches such as variational frameworks, or architectural extensions such as the Randomized Link (RL) Transformer. However, these approaches typically entail either additional difficulties during training/inference, or a significant increase in model size and complexity. Hence, we propose the \underline{Pa}rtially \underline{Ra}ndomized trans\underline{Former} (PaRaFormer), a simple extension of the transformer which involves freezing the weights of selected layers after random initialization. Experimental results reveal that the performance of the PaRaformer is comparable to that of the aforementioned approaches, despite not entailing any additional training difficulty or increase in model complexity.
\end{abstract}

\section{Introduction}
The development of open-domain dialogue agents, or chatbots, is a crucial objective in conversational AI. While advancements in deep learning and parallel computing have led to significant progress, a recurring challenge is the issue of low response diversity. This problem pertains to the agent's tendency to produce unvaried and repetitive responses, such as "That's fine" or "I'm not sure".

 Thus far, researchers have proposed numerous approaches to promoting response diversity. Recently, variational approaches in particular, have become extremely popular. These approaches involve incorporating variational frameworks such as the Conditional Variational Auto Encoder (CVAE) \cite{sun-etal-2021-generating,10.5555/3504035.3504704, gao-etal-2019-discrete,zhao-etal-2017-learning}, and Wasserstein Auto Encoder (WAE) \cite{https://doi.org/10.48550/arxiv.1711.01558} in an open-domain dialogue agent. Typically, a variational dialogue agent would consist of two additional networks responsible for generating the latent prior and approximated posterior distributions. During inference, a latent variable is randomly sampled from the latent prior distribution and passed to the decoder along with the dialogue context or dialogue history.Improvements in response diversity are attributed to the stochastic nature of sampling latent variables from the prior distribution. The agent is trained by minimizing the KL divergence or maximizing the evidence lower bound (ELBO) between the approximated posterior and the latent prior. However, variational dialogue agents face challenges such as the latent variable vanishing problem, which can be addressed with approaches like KL annealing, though it increases training difficulty. However, despite the increase in training difficulty and model complexity, CVAE-based frameworks have been employed in multiple controllable open-domain dialogue sub-tasks such as personalized \cite{icaart22, Song2019ExploitingPI, wu-etal-2020-guiding}, empathetic \cite{Ruan_2021,10.1145/3481890}, knowledge-based \cite{Wang_Si_Lei_Yang_2020} dialogue generation. Decoding strategies known to enhance diversity such as beam-search, temperature scaling, top-p/top-k sampling, also involve a trade-off with other aspects of dialogue quality such as coherence \cite{tevet-berant-2021-evaluating, 10.1162/tacl_a_00502}. Other prior approaches proposed to improve dialogue diversity also typically involve greater difficulty during either preprocessing, training, or inference (Section 2.1).

Due to the significant additional difficulty incurred by the aforementioned approaches, the Randomized Link (RL) Transformer \cite{lee-etal-2022-randomized}, an extension of the standard transformer \cite{NIPS2017_3f5ee243}, was recently proposed as an alternative. The RL Transformer successfully addresses the issue of low response diversity by introducing additional randomized layers to the standard transformer encoder and decoder architecture. During inference, the weights of these additional layers are frozen after random initialization. Stochasticity is induced via the additional randomized layers, which are randomly reinitialized every time a new dialogue context is presented to the model during inference. The responses generated by the RL Transformer showed comparable diversity to those of variational frameworks. Despite posing no extra training difficulty, the RL Transformer exhibits a significant increase in the number of parameters. This negatively affects scalability due to the additional randomized layers, each containing a relatively large number of neurons. A detailed comparison is available in Appendix A.1.



Hence, we propose the \underline{Pa}rtially \underline{Ra}ndomized trans\underline{Former} (PaRaFormer), an extension of the transformer which promotes response diversity by \emph{appropriately} initializing and freezing the weights of \emph{selected} layers in the transformer. Essentially, the weights of specific layers in the self attention and feed forward component of a transformer are frozen after initialization. 
During training, we adjust the variance of the weight initialization function to attain the maximum level of response diversification without compromising on other aspects of dialogue quality. 
Unlike prior approaches to promoting response diversity, the PaRaFormer does not entail any additional training difficulty or any increase in model size. 
Similar to variational frameworks, PaRaFormer improves response diversity by introducing stochasticity during response generation. However, like the RL Transformer, instead of random sampling, stochasticity is introduced via random weight initialization. Empirical results reveal that the PaRaformer is capable of generating contextually coherent responses that are comparable to responses generated by the RL Transformer as well as other variational frameworks in terms of response diversity. 


\section{Related Work}
\label{sec:2}
In addition to variational approaches, prior works have addressed the issue of low response diversity primarily via either altering the objective function, training target, or by utilizing alternative learning frameworks. These approaches, however, typically entail added difficulty during training. Several approaches propose novel objective functions aimed at promoting response diversity such as the Maximum Mutual Information (MMI) \cite{li-etal-2016-diversity}, the Inverse N-gram Frequency (INF) \cite{ueyama-kano-2020-diverse}, and the Frequency-Aware Cross-Entropy (FACE) \cite{10.1145/3308558.3313415}. Some approaches, on the other hand, introduce auxiliary loss terms alongside the standard MLE objective \cite{li-etal-2020-dont}. These new objective functions are typically significantly more complicated to evaluate. Other approaches such as label smoothing \cite{wang-etal-2021-diversifying} or softmax decomposition \cite{choi-etal-2020-f} involve actively modifying the training target. Adversarial dialogue generation frameworks which involve training additional discriminator networks \cite{holtzman-etal-2018-learning, li-etal-2017-adversarial}, and reinforcement learning approaches which entail defining a separate reward generation framework/model \cite{DBLP:journals/corr/abs-2111-13833, DBLP:journals/corr/abs-1811-05696} have also been proposed. 

Numerous randomization-based neural network architectures have also been proposed. Single-layer feed forward neural networks featuring randomly initialized, frozen weights such as the Extreme Learning Machine (ELM) \cite{1380068} and Random Vector Functional Link network \cite{144401}, have been shown to retain the universal approximation qualities of a fully trainable network \cite{DBLP:journals/corr/abs-2007-15776, 10.1109/TNN.2006.875977}. More recently, multiple deep variants of these approaches have also been introduced \cite{SHI2021107978, https://doi.org/10.48550/arxiv.2101.10265}. In the context of recurrent networks, researchers have proposed randomization-based architectures such as Echo State Networks \cite{articleEcho}, Liquid State Machines \cite{MAASS2004593}, and reservoir computing \cite{LUKOSEVICIUS2009127} networks have been introduced. Randomization-based convolutional networks \cite{https://doi.org/10.48550/arxiv.2007.13003} have also been introduced. For transformer models, \cite{https://doi.org/10.48550/arxiv.2005.00743} and \cite{shen-etal-2021-reservoir} have introduced randomized variants which achieved improved performance on several language modeling and machine translation tasks.

\section{PaRaFormer}

Generating open-domain dialogue involves generating a response $Y$ based on the dialogue context or dialogue history $X$. The response label is denoted by $\bar{Y}$, and $N$ refers to the number of encoder and decoder components. The PaRaFormer consists of regular transformer encoders and decoders interspersed between partially randomized (PaRa) encoder or a partially randomized (PaRa) decoder respectively. We chose to alternate between a PaRa encoder/decoder and a standard (fully-trainable) encoder/decoder as consecutive PaRa encoders/decoders would negatively impact the model's learning ability. This would result in a drop in response quality primarily in terms of coherence (Section 5.3). An illustration is provided in Figure 1. The PaRa encoder consists of PaRa attention network and a PaRa feed forward network. The PaRa decoder consists of a PaRa attention network, followed by a regular attention network, and a PaRa feed forward network. We found that replacing both attention networks in the decoder with PaRa attention networks would prevent the model form converging. An overview is provided in Figure 2.

Randomly initializing selected layers would enhance response diversity by introducing stochasticity or `randomness' during response generation. This will prevent the open-domain dialogue agent from defaulting to generic, repetitive responses. During training, frozen layers in the PaRaformer are reinitialized every epoch to allow the model to adapt to randomized weights, and generate diverse responses while maintaining overall quality. 

\begin{figure}
    \centering
    \includegraphics[width=70mm, height=35mm]{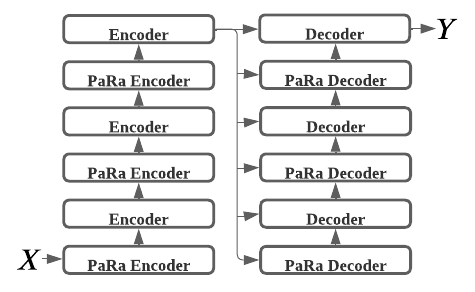}
    \caption{Overview of the PaRaFormer where $N = 6$.}
    \label{fig:my_label}
\end{figure}
\begin{figure}
    \centering
    \includegraphics[width=70mm, height=40mm]{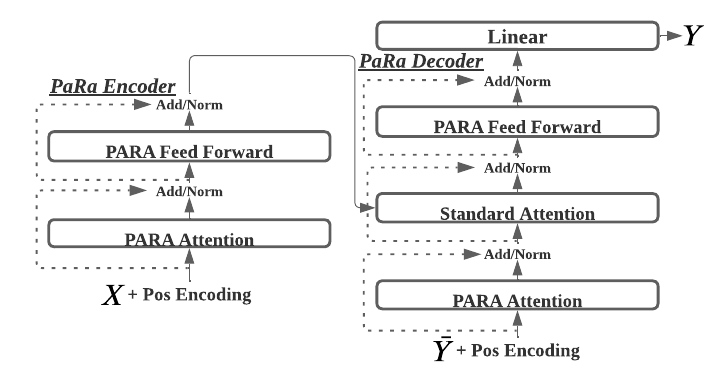}
    \caption{Overview of the PaRa encoder and decoder.}
    \label{fig:my_label}
\end{figure}

\subsection{PaRa Attention} 

To attain the query ($Q$), key ($K$), and value ($V$) vectors, the dialogue context $X$ is fed to three distinct linear layers with randomly initialized, frozen weights, denoted by $\textbf{W}^{r}_{Q}$, $\textbf{W}^{r}_{K}$, and $\textbf{W}^{r}_{V}$ respectively:
\begin{equation}
    Q = \textbf{W}^{r}_{Q}(X)
\end{equation}
\begin{equation}
    K = \textbf{W}^{r}_{K}(X)
\end{equation}
\begin{equation}
    V = \textbf{W}^{r}_{V}(X)
\end{equation}
where the superscript $r$ indicates a randomly initialized, frozen linear layer. $d_{Q}$, $d_{K}$, and $d_{V}$ refer to the dimensions of $\textbf{W}^{r}_{Q}$, $\textbf{W}^{r}_{K}$, and $\textbf{W}^{r}_{V}$ respectively. $n$ refers to the embedding size. It should be noted that the input to the standard attention network in the decoder consists of the output of the encoders and the output of the prior PaRa decoder. Subsequently, the dot product of the $Q$ and $K$ vectors is computed and divided by the square root of the size of $Q$ and $K$, which is denoted by $d^{k}$. Then, we apply the softmax function to the computed score and multiply with the $V$ vector to attain the output of the PaRa attention network, denoted by $Z$:
\begin{equation}
    Z = Softmax(\frac{QK^{T}}{\sqrt{d_{k}}})V
\end{equation}
where $T$ refers to the transpose operation. Finally, to obtain the output of the PaRa Attention network, $Z$ is passed to a single trainable linear layer $\textbf{W}^{r}_{Z}$:
\begin{equation}
    attn\_out = \textbf{W}_{Z}(Z)
\end{equation}
where $attn\_out$ refers to output of the PaRa attention network. We found that replacing $\textbf{W}_{Z}$ with a randomly initialized, frozen layer would degrade overall response quality. An overview is provided in Figure 3(a).

Similar to the regular transformer, the multi-headed variant of the PaRa attention network involves defining multiple parallel PaRa attention networks. The output from each network is concatenated and passed to the subsequent PaRa feed forward network.

\begin{figure*}
    \centering
    \includegraphics[width=120mm,height=35mm]{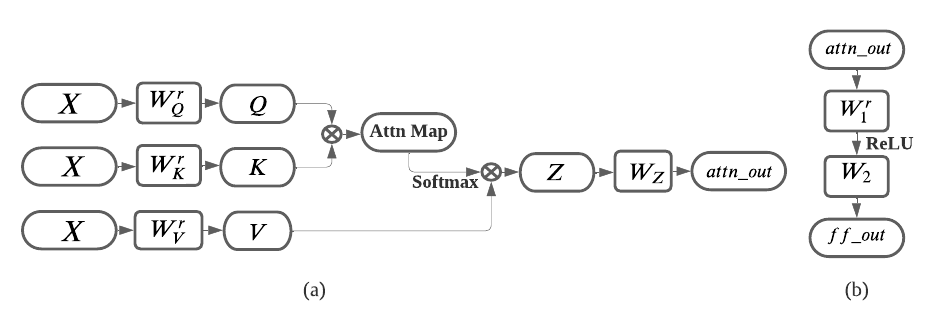}
    \caption{(a) Overview of the PaRa attention network. (b) Overview of the PaRa feed forward network.}
    \label{fig:my_label}
\end{figure*}

\subsection{PaRa Feed Forward} 
The input to the PaRa Feed Forward network is first fed to a linear layer with randomly initialized frozen weights and biases denoted by $\textbf{W}_{1}^{r}$ and $\textbf{b}_{1}^{r}$. Then, the ReLU activation function is applied. The resultant output is then passed to a trainable layer denoted by $\textbf{W}_{2}$ and $\textbf{b}_{2}$:
\begin{equation}
    ff\_out = \textbf{W}_{2}(ReLU(\textbf{W}^{r}_{1}(attn\_out)+\textbf{b}_{1}^{r}))+\textbf{b}_{2}
\end{equation}
where $ff\_out$ refers to the output of the PaRa feed forward network. $d_{1}$ and $n$ refer to the size of feed-forward layers $\textbf{W}_{1}^{r}$ and $\textbf{W}_{2}$ respectively. An overview is provided in Figure 3(b).

\subsection{Random Weight Initialization}
During training, the frozen weights are reinitialized every epoch. Hence, the weight initialization function would directly impact PaRaFormer's ability to learn and the amount of stochasticity induced during generation, which is directly proportional to response diversity. A weight initialization function with an excessively large variance increases stochasticity and diversity, but at the expense of PaRaFormer's ability to learn. Unlike the RL Transformer, in this work, we adjust the variance of weight initialization in the randomized layers to attain maximum response diversification without compromising on other aspects of response quality such as coherence. In our experimentation, we consider two different weight initialization functions: the standard Normal initialization and the Scalable Kaiming initialization, which is an extension of the Kaiming Normal initialization.

\noindent\textbf{Standard Normal Initialization.} During training, the weight vectors $\textbf{W}^{r}_{Q}$, $\textbf{W}^{r}_{K}$, $\textbf{W}^{r}_{V}$, $\textbf{W}_{1}^{r}$ and bias vector $\textbf{b}_{1}^{r}$ are randomly initialized every epoch via Standard Normal initialization:
\begin{equation}
    \textbf{W}_{Q}^{r}, \textbf{W}_{K}^{r}, \textbf{W}_{V}^{r} \sim N(0.0,\sigma_{SA}^{2})
\end{equation}
\begin{equation}
    \textbf{W}_{1}^{r}, \textbf{b}_{1}^{r} \sim N(0.0,\sigma_{FF}^{2})
\end{equation}
where $\sigma_{SA}$ and $\sigma_{FF}$ refer to the standard deviation utilized during random weight initialization in the PaRa self attention and PaRa feed forward networks respectively. Both $\sigma_{SA}$ and $\sigma_{FF}$ are regarded as additional hyperparameters to be tuned during training. 

\noindent\textbf{Scalable Kaiming Initialization.} Similar to Xavier initialization \cite{pmlr-v9-glorot10a}, the Kaiming Normal initialization \cite{7410480} ensures that the variance of all layers in a neural network are equal. This prevents the vanishing and exploding gradient problem by ensuring that the layer outputs are not too small or too large respectively. However, unlike the Xavier initialization, the Kaiming Normal initialization accounts for the activation function applied to the layer input. The activation function applied to the layer input is considered instead of the activation function applied to the output as we will be utilizing the forward pass variant of the Kaiming Normal initialization. \cite{7410480} showed that if the ReLU activation is applied to the layer input, the weight initialization should be constrained by $\frac{1}{2}n_{i}Var(\textbf{W}_{i}) = 1$, where $n$ refers to the number of layer inputs and $i$ refers to an arbitrary layer in the network. This results in  a standard deviation of $\frac{\sqrt{2}}{\sqrt{n_{i}}}$ i.e., $\textbf{W} \sim N(0,\frac{2}{n_{i}})$.

However, in the PaRa attention network, since the randomized layers $\textbf{W}_{Q}^{r}, \textbf{W}_{K}^{r}, \textbf{W}_{V}^{r}$, are used to generate $Q$, $K$, and $V$ respectively, no activation function is applied to the layer inputs. Also in the PaRa feed forward network, $\textbf{W}_{1}^{r}$ and $\textbf{b}_{1}^{r}$ precedes the ReLU activation. Hence, weight initialization should be constrained by $n_{i}Var(\textbf{W}_{i}) = 1$ instead, resulting in a standard deviation of $\frac{1}{\sqrt{n_{i}}}$ i.e., $\textbf{W} \sim N(0,\frac{1}{n_{i}})$ (shown in Appendix A.2).

Since the Kaiming Normal initialization was designed to prevent the exploding and vanishing gradient problem, the standard deviation value used would neither result in a complete degradation of the model's learning ability nor a complete lack of stochasticity. Thus, the standard deviation value used in the Kaiming Normal initialization would be a suitable base value from which further scaling can be introduced. Hence, we introduce a scalable Kaiming initialization for random weight initialization. We hypothesize that scaling the standard deviation value used in the Kaiming Normal initialization would allow for further response diversification without negatively impacting the model's learning ability. In our implementation, we utilize gain parameters to scale the variance of initialization. This allows us to manually tune the amount of stochasticity induced in the generation process, further diversifying the generated responses. This results in the following weight initializations:
\begin{equation}
    \textbf{W}_{Q}^{r}, \textbf{W}_{K}^{r}, \textbf{W}_{V}^{r} \sim N(0.0,\frac{\gamma_{SA}^{2}}{n_{i}})
\end{equation}
\begin{equation}
   \textbf{W}_{1}^{r}, \textbf{b}_{1}^{r} \sim N(0.0,\frac{\gamma_{FF}^{2}}{n_{i}})
\end{equation}
where $\gamma_{SA}$ and $\gamma_{FF}$ refer to the gain parameters, and $\frac{\gamma_{SA}}{\sqrt{n_{i}}}$ and $\frac{\gamma_{FF}}{\sqrt{n_{i}}}$ represent the standard deviations of the random weight initialization in the PaRa self attention and PaRa feed forward networks respectively.

\section{Experiment}

\textbf{Corpora} In our experiments, we use two main datasets: DailyDialog \cite{li-etal-2017-dailydialog} and EmpatheticDialogues \cite{rashkin-etal-2019-towards}. The DailyDialog corpus contains diverse, open-domain multi-turn conversations covering various styles, emotions, and topics. On the other hand, the EmpatheticDialogues dataset is designed to train and evaluate dialogue systems on their empathetic responses. It consists of pairs of conversations, where one speaker shares an event, and the other responds empathetically. In our experiments, the dialogue agent's task is to generate responses based solely on the context of the ongoing conversation. We do not use any additional information such as response labels (e.g., emotion, topic, or style) or speaker labels. The dialogue context comprises a maximum of 5 dialogue turns.

\noindent\textbf{Implementation} For our implementation, the PaRaFormer consists of six encoders and decoders, with four attention heads. Since the 300d GloVe embedding \cite{pennington2014glove} is used, the embedding size $n = 300$. $d_{k}$, $d_{v}$ and $d_{z}$ are fixed at 128. $d_{ff}$ is set to 2048. During training, the Adam optimizer (learning rate = 0.0006, batch size = 32) is used. In our experiments, most responses are generated via greedy decoding. We utilize greedy decoding instead of beam-search or other sampling-based decoding methods so that any gains in diversity can be attributed directly to the model architecture.

\noindent\textbf{Baselines} For our experiments, we implement two variants of the PaRaFormer: $PaRaFormer_{N}$ and $PaRaFormer_{K}$. For random weight initialization, $PaRaFormer_{N}$ utilizes Standard Normal initialization ($\sigma_{SA} = 0.01$, $\sigma_{FF}=0.05$) and $PaRaFormer_{K}$ employs Kaiming Normal initialization ($\gamma_{SA} = 2.5$, $\gamma_{FF}=1.5$). We implement three encoder-decoder models: a standard Transformer $Transformer$ \cite{NIPS2017_3f5ee243}; a $Seq2seq$ model with attention \cite{https://doi.org/10.48550/arxiv.1409.0473}; a Hierarchical Recurrent Encoder Decoder ($HRED$) \cite{10.5555/3016387.3016435}. Additionally, due to the success and popularity of variational frameworks in recent years, we implement four variational models: a Variational Hierarchical Recurrent Encoder Decoder ($VHRED$) \cite{Serban_Sordoni_Lowe_Charlin_Pineau_Courville_Bengio_2017}; a Variational Hierarchical Conversation RNNs ($VHCR$) \cite{park-etal-2018-hierarchical}; a transformer-based $CVAE$ \cite{zhao-etal-2017-learning} (section 3.2); and the Sequential Variational Transformer ($SVT$) \cite{https://doi.org/10.48550/arxiv.2003.12738}, which features a variational decoder that implicitly generates a distinct latent variable for each position. For $Seq2seq$, $HRED$, $VHRED$ and $VHCR$, all encoder and decoder components consist of two GRUs (hidden\_dim = 512). For all variational models, the prior and approximated posterior distributions are defined by MLPs (num\_layers=3, hidden\_dim=512, latent\_dim=300). The variational transformer baselines ($CVAE$, and $SVT$) consist of six encoder and decoders, and four attention heads (identical to the PaRaFormer). In addition, we also implement the $RL\:Transformer$ \cite{lee-etal-2022-randomized}. All transformer-based baselines ($Transformer$, $CVAE$, $SVT$, and $RL\:Transformer$) consist of six encoder and decoders, and four attention heads (identical to the PaRaFormer). This would ensure that any improvements in performance can be attributed to our proposed architectural enhancements instead of model size. Additionally, we benchmark our baselines against the standard GPT-2 pretrained language model ($GPT-2$), which was finetuned on the DailyDialog corpus. Due to computational constraints, we utilize the small GPT-2 model from HuggingFace (12 decoders). Responses are generated via greedy decoding. 

\noindent\textbf{Automatic Evaluation} To quantify diversity, we utilize the Distinct-$n$ metric ($n=1,2,3$) \cite{li-etal-2016-diversity}, which quantifies the number of distinct $n$-grams in the generated responses. A higher distinct score is indicative of greater overall response diversity. We do not rely on metrics drawn from machine translation such as ROUGE \cite{lin-2004-rouge} and METEOR \cite{banerjee-lavie-2005-meteor}, which involves comparing the generated response to the reference response, as prior work have shown that these metrics are extremely poor at measuring the quality of a generated response and do not correspond to any aspect of human evaluation \cite{liu-etal-2016-evaluate}. To measure the contextual coherence of the generated response, we utilize the Utterance Entailment (UE) score \cite{9747458}. Essentially, computing the UE score involves applying a BERT-based Natural Language Inference model to the generated response and each utterance in the dialogue context.

\noindent\textbf{Human Evaluation} In addition, we also employ human evaluation. We engaged five participants with high levels of English proficiency to evaluate the responses generated by the PaRaFormer against the other implemented baselines based on `Diversity', `Fluency', and `Coherence'. `Diversity' refers to the overall variability of the generated responses in terms of vocabulary, `Fluency' encompasses the eloquence of the responses, and `Coherence' quantifies contextual coherence i.e., the propriety/suitability of the generated response with regard to the dialogue context. Each participant evaluated 50 randomly selected dialogue examples, comparing PaRaFormer's response with other baselines without knowing the generating model. For each criteria, each participant was told to evaluate if the response generated by the PaRaFormer variant either wins, loses, or ties with the response generated by the other baselines. The win, loss, and tie rates for each comparison is provided in Table 3.

\section{Results \& Discussion}

\begin{table}
\caption{Overview of the automatic evaluation results on the DailyDialog corpus. * indicates statistically significant
differences (t-test, $p$-value \textless 0.01) from the best result in that column (\textbf{bolded}).}
\centering
\scalebox{0.65}{
\begin{tabular}{ccccc} 
\hline
                               & Dist-1               & Dist-2               & Dist-3               & UE                    \\ 
\hline
$Seq2seq$                      & 0.005*                & 0.017*                & 0.032*                &0.051*           \\
$HRED$                         & 0.012*                & 0.063*                & 0.141*                &0.073*                  \\
$VHRED$                        & 0.014*                & 0.131*                & 0.262*                &0.063*                  \\
$VHCR$                         & 0.010*                & 0.073*                & 0.186*                &0.062*                  \\
$Transformer$                  & 0.011*                & 0.106*                & 0.168*                &0.076                  \\
$CVAE$                         & 0.040*                & 0.183*                & 0.446                &0.061*                 \\
$SVT$                          & 0.037*                & 0.169*                & 0.441                &0.063*                  \\
$GPT-2$                         & 0.017*                & 0.174*                & 0.368*                &0.081                 \\
$RL\:Transformer$               &0.045                & 0.216               &0.444                &0.069*                  \\
\hline
\multicolumn{1}{l}{$PaRaFormer_K$} & \multicolumn{1}{c}{\textbf{0.051}} & \multicolumn{1}{c}{\textbf{0.236}} & \multicolumn{1}{c}{\textbf{0.467}} & \multicolumn{1}{c}{0.082}  \\
\multicolumn{1}{l}{$PaRaFormer_N$} & \multicolumn{1}{c}{0.039*} & \multicolumn{1}{c}{0.193} & \multicolumn{1}{c}{0.428} & \textbf{0.085}                      \\
\hline
\end{tabular}}
\end{table}

\begin{table}
\caption{Overview of the automatic evaluation results on the EmpatheticDialogues corpus. * indicates statistically significant
differences (t-test, $p$-value \textless 0.01) from the best result in that column (\textbf{bolded}).}
\centering
\scalebox{0.65}{
\begin{tabular}{ccccc} 
\hline
                               & Dist-1               & Dist-2               & Dist-3               & UE                    \\ 
\hline
$Seq2seq$                      & 0.002*                & 0.009*                & 0.189*                &0.021*           \\
$HRED$                         & 0.017*                & 0.044*                & 0.225*                &0.043*                  \\
$VHRED$                        & 0.028*                & 0.174*                & 0.301*                &0.034*                  \\
$VHCR$                         & 0.026*                & 0.123*                & 0.253*                &0.041*                  \\
$Transformer$                  & 0.023*                & 0.117*                & 0.221*                &0.053*                  \\
$CVAE$                         & 0.031*                & 0.226*                & 0.426                 &0.051*                 \\
$SVT$                          & 0.025*                & 0.251*                & 0.484                 &0.054*                  \\
$GPT-2$                         & 0.027*                & 0.134*                &0.392*                &0.071*                  \\
$RL\:Transformer$               &0.036                &0.265             &0.509                &0.063*                  \\
\hline
\multicolumn{1}{l}{$PaRaFormer_K$} & \multicolumn{1}{l}{\textbf{0.043}} & \multicolumn{1}{c}{\textbf{0.288}} & \multicolumn{1}{c}{\textbf{0.521}} & \multicolumn{1}{c}{\textbf{0.079}}  \\
\multicolumn{1}{l}{$PaRaFormer_N$} & \multicolumn{1}{c}{0.038} & \multicolumn{1}{c}{0.273} & \multicolumn{1}{c}{0.488} & 0.077                     \\
\hline
\end{tabular}}
\end{table}

\begin{table*}
\caption{Human evaluation results on the DailyDialog corpus. Kappa values \cite{fleiss1971mns}, represented by $\kappa$, typically range from 0.4 to 0.6, indicating moderate inter-rater agreement.}
\centering
\scalebox{0.65}{
\begin{tabular}{ccccccccccccc} 
\hline
                  & \multicolumn{4}{c}{Fluency} & \multicolumn{4}{c}{Diversity} & \multicolumn{4}{c}{Coherence}  \\ 
\cline{2-13}
                  & Win  & Tie  & Loss & $\kappa$  & Win  & Tie  & Loss & $\kappa$    & Win  & Tie  & Loss & $\kappa$     \\ 
\hline
$PaRaFormer_{N}$ vs $Seq2seq$         & 37\% & 39\% & 24\% & 0.44   & 70\% & 19\% & 11\% & 0.47     & 40\% & 45\% & 15\% & 0.55      \\
$PaRaFormer_{N}$ vs $HRED$            & 39\% & 31\% & 30\% & 0.48   & 68\% & 22\% & 10\% & 0.44     & 45\% & 39\% & 16\% & 0.61      \\
$PaRaFormer_{N}$ vs $VHRED$           & 47\% & 32\% & 21\% & 0.39   & 58\% & 34\% & 18\% & 0.50     & 41\% & 40\% & 19\% & 0.42      \\
$PaRaFormer_{N}$ vs $VHCR$            & 42\% & 29\% & 27\%  & 0.59  & 61\% & 35\% & 14\% & 0.59     & 44\% & 40\% & 16\%  & 0.49      \\
$PaRaFormer_{N}$ vs $Transformer$     & 36\% & 41\% & 22\% & 0.46   & 67\% & 28\% & 5\%  & 0.57     & 46\% & 36\% & 18\% & 0.56      \\
$PaRaFormer_{N}$ vs $CVAE$            & 41\% & 34\% & 25\% & 0.45   & 45\% & 38\% & 17\% & 0.53     & 41\% & 36\% & 23\% & 0.55      \\
$PaRaFormer_{N}$ vs $SVT$             & 45\% & 37\% & 18\%  & 0.49  & 47\% & 39\% & 14\% & 0.51     & 43\% & 39\% & 18\% & 0.51      \\
$PaRaFormer_{N}$ vs $GPT-2$           & 27\% & 48\% & 25\% & 0.53   & 49\% & 40\% & 11\% & 0.61     & 39\% & 38\% & 23\% & 0.53      \\
$PaRaFormer_{N}$ vs $RL\:Transformer$ & 36\% & 35\% & 29\% & 0.55   & 33\% & 46\% & 21\% & 0.48     & 36\% & 38\% & 26\% & 0.55      \\
\hline
 $PaRaFormer_{K}$ vs $Seq2seq$        & 35\% & 42\% & 23\% & 0.46   & 73\% & 21\% & 6\%  & 0.55     & 41\% & 43\% & 16\% & 0.48      \\
 $PaRaFormer_{K}$ vs $HRED$           & 37\% & 39\% & 24\% & 0.51   & 69\% & 20\% & 11\%  & 0.62    & 50\% & 41\% & 9\% & 0.50      \\
 $PaRaFormer_{K}$ vs $VHRED$          & 35\% & 36\% & 29\%  & 0.49  & 56\% & 38\% & 16\% & 0.59     & 51\% & 42\% & 7\% & 0.63      \\
 $PaRaFormer_{K}$ vs $VHCR$           & 39\% & 34\% & 27\%  & 0.55  & 64\% & 36\% & 10\% & 0.48     & 53\% & 41\% & 6\% & 0.59      \\
 $PaRaFormer_{K}$ vs $Transformer$    & 38\% & 39\% & 23\% & 0.53   & 67\% & 28\% & 5\% & 0.47     & 48\% & 44\% & 8\% & 0.49      \\
 $PaRaFormer_{K}$ vs $CVAE$           & 44\% & 33\% & 23\% & 0.52   & 49\% & 44\% & 7\% & 0.53     & 42\% & 42\% & 16\% & 0.52      \\
 $PaRaFormer_{K}$ vs $SVT$            & 49\% & 30\% & 21\% & 0.46   & 48\% & 40\% & 10\% & 0.51    & 43\% & 39\% & 18\% & 0.43      \\
 $PaRaFormer_{K}$ vs $GPT-2$          & 31\% & 45\% & 24\% & 0.45   & 50\% & 43\% & 7\% & 0.56     & 41\% & 34\% & 25\% & 0.47       \\
 $PaRaFormer_{K}$ vs $RL\:Transformer$ & 33\% & 41\% & 36\% & 0.49  & 42\% & 35\% & 23\% & 0.41     & 39\% & 40\% & 21\% & 0.57      \\
 \hline
 $PaRaFormer_{K}$ vs $PaRaFormer_{N}$  & 32\% & 37\% & 31\% & 0.57   & 31\% & 51\% & 18\% & 0.61     & 28\% & 39\% & 33\% & 0.54      \\
\hline
\end{tabular}}
\end{table*}

\begin{table}
\centering
\caption{Distinct-$n$  and UE scores for various PaRa encoder/decoder configurations. $PaRaFormer_{N}$ ($\sigma_{SA} = 0.01$, $\sigma_{FF}=0.05$) is used as the base model. * indicates statistically significant differences (t-test, $p$-value \textless 0.05) from the best result in that column (\textbf{bolded}).}
\scalebox{0.65}{
\begin{tabular}{ccccl} 
\hline
                 & Dist-1 & Dist-2 & Dist-3 & UE  \\ 
\hline
Alt & \textbf{0.039}  & \textbf{0.193}  & \textbf{0.428}  &\textbf{0.085}     \\
Full             & 0.018*  & 0.071*  & 0.183*  &0.047*     \\
Seq\_1           & 0.033  & 0.158*  & 0.357*  &0.059*     \\
Seq\_2           & 0.032*  & 0.179  & 0.407  &0.061*     \\
\hline
\end{tabular}}
\end{table}

\begin{table}
\caption{Distinct-$n$ and UE scores for $\sigma_{SA}, \sigma_{FF} = 0.01,0.05,0.50$.}
\centering
\scalebox{0.65}{
\begin{tabular}{cccccc} 
\hline
$\sigma_{SA}$   & $\sigma_{FF}$ & Dist-1 & Dist-2 & Dist-3 & UE  \\ 
\hline
0.01 &0.01&0.026*   &0.149*   &0.350*   &0.069*\\
0.01 &0.05&\textbf{0.039}   &\textbf{0.193}   &0.428   &\textbf{0.085}\\
0.01 &0.50&0.030   &0.112*   &0.230*   &0.027*\\
0.05 &0.01&0.024*   &0.129*   &0.311*   &0.070\\
0.05 &0.05&0.036   &0.159*   &\textbf{0.444}   &0.080\\
0.05 &0.50&0.027*   &0.111*   &0.358*   &0.023*\\
0.50 &0.01&0.005*   &0.023*   &0.055*   &0.020*\\
0.50 &0.05&0.006*   &0.024*   &0.057*   &0.019*\\
0.50 &0.50&0.001*   &0.017*   &0.041*   &0.013*\\
\hline
\end{tabular}}
\end{table}

\begin{table}
\caption{Distinct-$n$ and UE scores for $\gamma_{SA} = 1.5,2.5,3.5$ and $\gamma_{FF} = 1.0,1.5,2.0$.}
\centering
\scalebox{0.65}{
\begin{tabular}{cccccc}
\hline
$\gamma_{SA}$  & $\gamma_{FF}$  & Dist-1 & Dist-2 & Dist-3 & UE  \\ 
\hline
1.5 & 1.5 &0.031*   &0.143*   &0.320*   &0.074\\
1.5 & 2.5 &0.028*   &0.134*   &0.304*   &0.066*\\
1.5 & 3.5 &0.028*   &0.128*   &0.289*   &0.058*\\
2.5 & 1.5 &0.051   &\textbf{0.236}   &\textbf{0.467}   &\textbf{0.082}\\
2.5 & 2.5 &\textbf{0.053}   &0.194   &0.398   &0.057*\\
2.5 & 3.5 &0.039*  &0.173*   &0.373*   &0.055*\\
3.5 & 1.5 &0.033*   &0.131*   &0.282*   &0.040*\\
3.5 & 2.5 &0.033*   &0.126*   &0.267*   &0.044*\\
3.5 & 3.5 &0.011*   &0.054*   &0.126*   &0.029*\\
\hline
\end{tabular}}
\end{table}

\begin{table}
\centering
\caption{Distinct-$n$  and UE scores for various decoding strategies (applied to Transformer on DailyDialog). * indicates statistically significant differences (t-test, $p$-value \textless 0.05) from the best result in that column (\textbf{bolded}).}
\scalebox{0.7}{
\begin{tabular}{lcccl} 
\hline
                 & Dist-1 & Dist-2 & Dist-3 & UE  \\ 
\hline
$PaRaFormer_{K}$ & 0.051  & 0.236  & \textbf{0.467}  &0.082     \\
$PaRaFormer_{N}$ & 0.043*  & 0.193  & 0.428  &\textbf{0.085}     \\
Transformer      & 0.011*  & 0.106*  & 0.168  &0.076     \\
-$T=0.50$        &0.023*  &0.195*  &0.331*  &0.068*    \\
-$T=0.75$        &0.037*   &0.228   &0.396*   &0.063*                             \\
-$T=1.0$         &\textbf{0.057}  &\textbf{0.259}   &0.451 &0.051*    \\
-Top-p(0.9)     &0.039*   &0.244   &0.421*   &0.070*                             \\
-Top-k(40)      &0.035*   &0.213*   &0.403*   &0.067*                             \\
-Beam(5)        &0.031*   &0.196*   &0.358*   &0.063*                             \\
\hline
\end{tabular}}
\end{table}

\subsection{Quantitative Analysis}
Based on the results presented in Table 1, 2 and 3, it is apparent that $PaRaFormer_{K}$ outperformed $PaRaFormer_{N}$ in terms of response diversity. $PaRaFormer_{K}$ attained higher distinct-1,2, and 3 scores, as well as a higher percentage of Wins and lower percentage of Losses on the `Diversity' criterion relative to $PaRaFormer_{N}$. However, both $PaRaFormer_{N}$ and $PaRaFormer_{K}$ showed similar performance in terms of general fluency and contextual coherence. This can be inferred from the comparable UE scores, as well as the relatively similar percentages of Wins, Ties, and Losses between both $PaRaFormer_{N}$ and  $PaRaFormer_{K}$ on the `Fluency' and `Coherence' criterion (Table 3). 

The response diversity of both $PaRaFormer_{N}$ and $PaRaFormer_{K}$ is comparable to that of the $RL:Transformer$. When compared to all other implemented baselines (except for the $RL:Transformer$), both $PaRaFormer_{N}$ and $PaRaFormer_{K}$ achieved noticeably higher distinct scores and a significant percentage of wins on the 'Diversity' criterion, indicating that they produce more diverse responses. However, it is worth noting that the level of diversification attained by $PaRaFormer_{N}$ is generally slightly lower than that of $PaRaFormer_{K}$ and $RL:Transformer$. $PaRaFormer_{N}$ attained scores similar to those of the variational baselines, as evidenced by the relatively similar distinct scores and the high percentage of Ties on the 'Diversity' criterion.

Regarding contextual coherence, both $PaRaFormer_{N}$ and $PaRaFormer_{K}$ performed better than all variational baselines, showing higher Wins and Ties in the 'Coherence' criterion. The poor contextual coherence of variational baselines may be attributed to random sampling. Random sampling can lead to latent variables deviating too far from the prior distribution mean, resulting in incoherent responses. In terms of contextual coherence, both $PaRaFormer_{N}$ and $PaRaFormer_{K}$ achieved comparable performance to $RL:Transformer$, which outperformed all other baselines (except $GPT-2$). This is expected as $GPT-2$ is pretrained on a large amount of textual data, giving it greater language understanding capabilities. Regarding overall fluency, both $PaRaFormer_{N}$ and $PaRaFormer_{K}$ received similar human evaluation scores compared to all other implemented baselines. This is evident from the relatively consistent Win, Tie, and Loss scores of both models on the Fluency criterion against all implemented baselines.

\subsection{Qualitative Analysis}
The qualitative analysis confirms our initial observations. Non-variational baselines, with the exception of $RL:Transformer$, $PaRaFormer_{N}$, and $PaRaFormer_{K}$, tend to produce less diverse responses compared to their variational counterparts. These non-variational responses often consist of short, repetitive, and generic phrases, such as "Ok sure" or "Great." On the other hand, responses generated by $PaRaFormer_{N}$ and $PaRaFormer_{K}$ are noticeably more diverse relative to non-variational baselines, featuring a relatively larger number of unique responses. 

In addition, responses generated by variational baselines as well as $PaRaFormer_{N}$ and $PaRaFormer_{K}$ displayed a larger variability in terms of vocabulary and phrasing. Also, the contextual coherence of responses from variational baselines is relatively poorer than that of non-variational baselines. Some responses generated by variational models are unrelated to the dialogue context or directly contradict it. Samples of dialogue responses are provided in Table 8 from Appendix A.3.

\subsection{PaRa Encoder/Decoder Configuration}
In Table 4, we examine the performance of $PaRaFormer_{N}$ ($\sigma_{SA} = 0.01$, $\sigma_{FF}=0.05$) where every encoder and decoder is replaced with their PaRa counterpart (Full), and two variants where only the first (Seq\textsubscript{1}) and last (Seq\textsubscript{2}) $N/2$ encoders/decoders are replaced with their PaRa counterparts respectively. We can observe that the Full variant experienced a sharp drop in diversity. This can be attributed to the insufficient number of trainable weights in the model which hinders the model from learning effectively. Thus, the model defaults to generating short, highly repetitive, incoherent responses, which translates to low distinct and UE scores. Additionally, we also observe that sequential configurations would achieve lower levels of response diversification and coherence. This implies that consecutive PaRa components would likewise cause a degradation in learning efficacy.

\subsection{Gain \& Standard Deviation}
Table 5 shows the results obtained using different standard deviation values, $\sigma_{SA},\sigma_{FF} = 0.01, 0.05, 0.5$, for Standard Normal initialization. Table 6 presents the results with various gain values, $\gamma_{SA}, \gamma_{FF} = 1.5, 2.5, 3.5$, for Scalable Kaiming initialization.

A larger standard deviation during random initialization implies higher stochasticity or randomness. For both Standard Normal and Kaiming Normal initializations, smaller values of $\sigma_{SA}$, $\sigma_{FF}$, $\gamma_{SA}$, and $\gamma_{FF}$ lead to slightly lower distinct scores due to reduced stochasticity, while higher values of these parameters result in a drop in UE score, indicating decreased contextual coherence and learning ability.

Notably, the values of $\sigma_{FF}$ and $\gamma_{FF}$ significantly impact the agent's learning ability. Larger values hinder learning and lead to low-quality, generic responses with poor diversity and coherence. In contrast, the model shows relatively less sensitivity to high values of $\sigma_{SA}$ and $\gamma_{SA}$, aligning with prior research highlighting the importance of the feed forward component for transformer performance. However, a sharp drop in distinct and UE scores is observed when $\sigma_{FF}=0.5$ and $\gamma_{FF}=3.5$ due to excessive stochasticity, resulting in ineffective learning and gibberish generation. Similarly, lower values of $\sigma_{SA}$ and $\gamma_{SA}$ (0.01 and 1.5) lead to slightly lower distinct scores but comparable UE scores.

\subsection{Decoding Strategies}
 Additionally, we also compare the responses generated by a fine-tuned Transformer with various decoding strategies including temperature scaling, top-p, top-k, and beam search to $PaRaFormer_{N}$ and $PaRaFormer_{K}$. Results are presented in Table 7. Based on the results, it is apparent that utilizing the aforementioned decoding strategies would improve response diversity. However, as evidenced by the decreasing UE scores, this is typically accompanied by a drop in coherence. $PaRaFormer_{N}$ and $PaRaFormer_{K}$, on the other hand, achieved high levels of response diversification while maintaining coherence.

\section{Conclusion}
This paper introduces PaRaFormer, a straightforward extension of the transformer that incorporates randomly initialized, frozen weights in specific linear layers. Experimental results demonstrate that PaRaFormer is capable of generating diverse responses without compromising on contextual coherence. Future research could focus on exploring randomization-based methods in large language models. Particularly, investigating the application of pretraining techniques to a substantially larger PaRaFormer model and benchmarking it against other pretrained language models fine-tuned for open-domain dialogue generation. Moreover, further exploration of alternative randomization methods, like monte carlo dropout during inference, could be considered. 

\section{Limitations}
A crucial constraint is that this framework cannot leverage existing pretrained language models. The utilization of PaRaFromer necessitates training a PaRaFormer model from scratch. Moreover, the scope of this study does not encompass controllable dialogue tasks, such as personalized or knowledge-grounded dialogue generation, which are essential for natural, human-like open-domain conversation. Further research could explore the performance of PaFaFormer on controllable generation tasks and investigate the potential effects of frozen, randomly initialized weights on controllability.

\bibliography{anthology,custom}

\begin{thebibliography}{52}
\expandafter\ifx\csname natexlab\endcsname\relax\def\natexlab#1{#1}\fi

\bibitem[{Altan and Kutlu(2021)}]{https://doi.org/10.48550/arxiv.2101.10265}
Gokhan Altan and Yakup Kutlu. 2021.
\newblock Superiorities of deep extreme learning machines against convolutional
  neural networks.

\bibitem[{Bahdanau et~al.(2014)Bahdanau, Cho, and
  Bengio}]{https://doi.org/10.48550/arxiv.1409.0473}
Dzmitry Bahdanau, Kyunghyun Cho, and Yoshua Bengio. 2014.
\newblock Neural machine translation by jointly learning to align and
  translate.

\bibitem[{Banerjee and Lavie(2005)}]{banerjee-lavie-2005-meteor}
Satanjeev Banerjee and Alon Lavie. 2005.
\newblock {METEOR}: An automatic metric for {MT} evaluation with improved
  correlation with human judgments.
\newblock In \emph{Proceedings of the {ACL} Workshop on Intrinsic and Extrinsic
  Evaluation Measures for Machine Translation and/or Summarization}, pages
  65--72, Ann Arbor, Michigan. Association for Computational Linguistics.

\bibitem[{Choi et~al.(2020)Choi, Hong, Park, and Lee}]{choi-etal-2020-f}
Byung-Ju Choi, Jimin Hong, David Park, and Sang~Wan Lee. 2020.
\newblock F{\^{}}2-softmax: Diversifying neural text generation via frequency
  factorized softmax.
\newblock In \emph{Proceedings of the 2020 Conference on Empirical Methods in
  Natural Language Processing (EMNLP)}, pages 9167--9182, Online. Association
  for Computational Linguistics.

\bibitem[{Fleiss et~al.(1971)}]{fleiss1971mns}
J.L. Fleiss et~al. 1971.
\newblock {Measuring nominal scale agreement among many raters}.
\newblock \emph{Psychological Bulletin}, 76(5):378--382.

\bibitem[{Gao et~al.(2018)Gao, Bi, Liu, Li, and
  Shi}]{DBLP:journals/corr/abs-1811-05696}
Jun Gao, Wei Bi, Xiaojiang Liu, Junhui Li, and Shuming Shi. 2018.
\newblock \href {http://arxiv.org/abs/1811.05696} {Generating multiple diverse
  responses for short-text conversation}.
\newblock \emph{CoRR}, abs/1811.05696.

\bibitem[{Gao et~al.(2019)Gao, Bi, Liu, Li, Zhou, and
  Shi}]{gao-etal-2019-discrete}
Jun Gao, Wei Bi, Xiaojiang Liu, Junhui Li, Guodong Zhou, and Shuming Shi. 2019.
\newblock A discrete {CVAE} for response generation on short-text conversation.
\newblock In \emph{Proceedings of the 2019 Conference on Empirical Methods in
  Natural Language Processing and the 9th International Joint Conference on
  Natural Language Processing (EMNLP-IJCNLP)}, pages 1898--1908, Hong Kong,
  China. Association for Computational Linguistics.

\bibitem[{Glorot and Bengio(2010)}]{pmlr-v9-glorot10a}
Xavier Glorot and Yoshua Bengio. 2010.
\newblock Understanding the difficulty of training deep feedforward neural
  networks.
\newblock In \emph{Proceedings of the Thirteenth International Conference on
  Artificial Intelligence and Statistics}, volume~9 of \emph{Proceedings of
  Machine Learning Research}, pages 249--256, Chia Laguna Resort, Sardinia,
  Italy. PMLR.

\bibitem[{He et~al.(2015)He, Zhang, Ren, and Sun}]{7410480}
Kaiming He, Xiangyu Zhang, Shaoqing Ren, and Jian Sun. 2015.
\newblock Delving deep into rectifiers: Surpassing human-level performance on
  imagenet classification.
\newblock In \emph{2015 IEEE International Conference on Computer Vision
  (ICCV)}, pages 1026--1034.

\bibitem[{Holtzman et~al.(2018)Holtzman, Buys, Forbes, Bosselut, Golub, and
  Choi}]{holtzman-etal-2018-learning}
Ari Holtzman, Jan Buys, Maxwell Forbes, Antoine Bosselut, David Golub, and
  Yejin Choi. 2018.
\newblock Learning to write with cooperative discriminators.
\newblock In \emph{Proceedings of the 56th Annual Meeting of the Association
  for Computational Linguistics (Volume 1: Long Papers)}, pages 1638--1649,
  Melbourne, Australia. Association for Computational Linguistics.

\bibitem[{Huang et~al.(2006)Huang, Chen, and Siew}]{10.1109/TNN.2006.875977}
Guang-Bin Huang, Lei Chen, and Chee-Kheong Siew. 2006.
\newblock Universal approximation using incremental constructive feedforward
  networks with random hidden nodes.
\newblock \emph{Trans. Neur. Netw.}, 17(4):879–892.

\bibitem[{Huang et~al.(2004)Huang, Zhu, and Siew}]{1380068}
Guang-Bin Huang, Qin-Yu Zhu, and Chee-Kheong Siew. 2004.
\newblock Extreme learning machine: a new learning scheme of feedforward neural
  networks.
\newblock In \emph{2004 IEEE International Joint Conference on Neural Networks
  (IEEE Cat. No.04CH37541)}, volume~2, pages 985--990 vol.2.

\bibitem[{Jaeger(2001)}]{articleEcho}
Herbert Jaeger. 2001.
\newblock The" echo state" approach to analysing and training recurrent neural
  networks-with an erratum note'.
\newblock \emph{Bonn, Germany: German National Research Center for Information
  Technology GMD Technical Report}, 148.

\bibitem[{Jiang et~al.(2019)Jiang, Ren, Monz, and
  de~Rijke}]{10.1145/3308558.3313415}
Shaojie Jiang, Pengjie Ren, Christof Monz, and Maarten de~Rijke. 2019.
\newblock Improving neural response diversity with frequency-aware
  cross-entropy loss.
\newblock In \emph{The World Wide Web Conference}, WWW '19, page 2879–2885,
  New York, NY, USA. Association for Computing Machinery.

\bibitem[{Lee et~al.(2022)Lee, Aik~Lee, and Gan}]{9747458}
Jing~Yang Lee, Kong Aik~Lee, and Woon~Seng Gan. 2022.
\newblock Improving contextual coherence in variational personalized and
  empathetic dialogue agents.
\newblock In \emph{ICASSP 2022 - 2022 IEEE International Conference on
  Acoustics, Speech and Signal Processing (ICASSP)}, pages 7052--7056.

\bibitem[{Lee. et~al.(2022)Lee., Lee., and Gan.}]{icaart22}
Jing~Yang Lee., Kong~Aik Lee., and Woon~Seng Gan. 2022.
\newblock Dlvgen: A dual latent variable approach to personalized dialogue
  generation.
\newblock In \emph{Proceedings of the 14th International Conference on Agents
  and Artificial Intelligence - Volume 2: ICAART}, pages 193--202. INSTICC,
  SciTePress.

\bibitem[{Lee et~al.(2022)Lee, Lee, and Gan}]{lee-etal-2022-randomized}
Jing~Yang Lee, Kong~Aik Lee, and Woon~Seng Gan. 2022.
\newblock A randomized link transformer for diverse open-domain dialogue
  generation.
\newblock In \emph{Proceedings of the 4th Workshop on NLP for Conversational
  AI}, pages 1--11, Dublin, Ireland. Association for Computational Linguistics.

\bibitem[{Li et~al.(2016)Li, Galley, Brockett, Gao, and
  Dolan}]{li-etal-2016-diversity}
Jiwei Li, Michel Galley, Chris Brockett, Jianfeng Gao, and Bill Dolan. 2016.
\newblock A diversity-promoting objective function for neural conversation
  models.
\newblock In \emph{Proceedings of the 2016 Conference of the North {A}merican
  Chapter of the Association for Computational Linguistics: Human Language
  Technologies}, pages 110--119, San Diego, California. Association for
  Computational Linguistics.

\bibitem[{Li et~al.(2017{\natexlab{a}})Li, Monroe, Shi, Jean, Ritter, and
  Jurafsky}]{li-etal-2017-adversarial}
Jiwei Li, Will Monroe, Tianlin Shi, S{\'e}bastien Jean, Alan Ritter, and Dan
  Jurafsky. 2017{\natexlab{a}}.
\newblock Adversarial learning for neural dialogue generation.
\newblock In \emph{Proceedings of the 2017 Conference on Empirical Methods in
  Natural Language Processing}, pages 2157--2169, Copenhagen, Denmark.
  Association for Computational Linguistics.

\bibitem[{Li et~al.(2020)Li, Roller, Kulikov, Welleck, Boureau, Cho, and
  Weston}]{li-etal-2020-dont}
Margaret Li, Stephen Roller, Ilia Kulikov, Sean Welleck, Y-Lan Boureau,
  Kyunghyun Cho, and Jason Weston. 2020.
\newblock Don{'}t say that! making inconsistent dialogue unlikely with
  unlikelihood training.
\newblock In \emph{Proceedings of the 58th Annual Meeting of the Association
  for Computational Linguistics}, pages 4715--4728, Online. Association for
  Computational Linguistics.

\bibitem[{Li et~al.(2021)Li, Zhang, Lu, and Zong}]{10.1145/3481890}
Mei Li, Jiajun Zhang, Xiang Lu, and Chengqing Zong. 2021.
\newblock Dual-view conditional variational auto-encoder for emotional dialogue
  generation.
\newblock \emph{ACM Trans. Asian Low-Resour. Lang. Inf. Process.}, 21(3).

\bibitem[{Li et~al.(2017{\natexlab{b}})Li, Su, Shen, Li, Cao, and
  Niu}]{li-etal-2017-dailydialog}
Yanran Li, Hui Su, Xiaoyu Shen, Wenjie Li, Ziqiang Cao, and Shuzi Niu.
  2017{\natexlab{b}}.
\newblock {D}aily{D}ialog: A manually labelled multi-turn dialogue dataset.
\newblock In \emph{Proceedings of the Eighth International Joint Conference on
  Natural Language Processing (Volume 1: Long Papers)}, pages 986--995, Taipei,
  Taiwan. Asian Federation of Natural Language Processing.

\bibitem[{Lin(2004)}]{lin-2004-rouge}
Chin-Yew Lin. 2004.
\newblock {ROUGE}: A package for automatic evaluation of summaries.
\newblock In \emph{Text Summarization Branches Out}, pages 74--81, Barcelona,
  Spain. Association for Computational Linguistics.

\bibitem[{Lin et~al.(2020)Lin, Winata, Xu, Liu, and
  Fung}]{https://doi.org/10.48550/arxiv.2003.12738}
Zhaojiang Lin, Genta~Indra Winata, Peng Xu, Zihan Liu, and Pascale Fung. 2020.
\newblock Variational transformers for diverse response generation.

\bibitem[{Liu et~al.(2016)Liu, Lowe, Serban, Noseworthy, Charlin, and
  Pineau}]{liu-etal-2016-evaluate}
Chia-Wei Liu, Ryan Lowe, Iulian Serban, Mike Noseworthy, Laurent Charlin, and
  Joelle Pineau. 2016.
\newblock How {NOT} to evaluate your dialogue system: An empirical study of
  unsupervised evaluation metrics for dialogue response generation.
\newblock In \emph{Proceedings of the 2016 Conference on Empirical Methods in
  Natural Language Processing}, pages 2122--2132, Austin, Texas. Association
  for Computational Linguistics.

\bibitem[{Lu et~al.(2021)Lu, Lam, Cheng, and
  Meng}]{DBLP:journals/corr/abs-2111-13833}
Hongyuan Lu, Wai Lam, Hong Cheng, and Helen~M. Meng. 2021.
\newblock \href {http://arxiv.org/abs/2111.13833} {Partner personas generation
  for diverse dialogue generation}.
\newblock \emph{CoRR}, abs/2111.13833.

\bibitem[{Lukoševičius and Jaeger(2009)}]{LUKOSEVICIUS2009127}
Mantas Lukoševičius and Herbert Jaeger. 2009.
\newblock Reservoir computing approaches to recurrent neural network training.
\newblock \emph{Computer Science Review}, 3(3):127--149.

\bibitem[{Maass and Markram(2004)}]{MAASS2004593}
Wolfgang Maass and Henry Markram. 2004.
\newblock On the computational power of circuits of spiking neurons.
\newblock \emph{Journal of Computer and System Sciences}, 69(4):593--616.

\bibitem[{Needell et~al.(2020)Needell, Nelson, Saab, and
  Salanevich}]{DBLP:journals/corr/abs-2007-15776}
Deanna Needell, Aaron~A. Nelson, Rayan Saab, and Palina Salanevich. 2020.
\newblock \href {http://arxiv.org/abs/2007.15776} {Random vector functional
  link networks for function approximation on manifolds}.
\newblock \emph{CoRR}, abs/2007.15776.

\bibitem[{Pao and Takefuji(1992)}]{144401}
Y.-H. Pao and Y.~Takefuji. 1992.
\newblock Functional-link net computing: theory, system architecture, and
  functionalities.
\newblock \emph{Computer}, 25(5):76--79.

\bibitem[{Park et~al.(2018)Park, Cho, and Kim}]{park-etal-2018-hierarchical}
Yookoon Park, Jaemin Cho, and Gunhee Kim. 2018.
\newblock A hierarchical latent structure for variational conversation
  modeling.
\newblock In \emph{Proceedings of the 2018 Conference of the North {A}merican
  Chapter of the Association for Computational Linguistics: Human Language
  Technologies, Volume 1 (Long Papers)}, pages 1792--1801, New Orleans,
  Louisiana. Association for Computational Linguistics.

\bibitem[{Pennington et~al.(2014)Pennington, Socher, and
  Manning}]{pennington2014glove}
Jeffrey Pennington, Richard Socher, and Christopher~D. Manning. 2014.
\newblock Glove: Global vectors for word representation.
\newblock In \emph{Empirical Methods in Natural Language Processing (EMNLP)},
  pages 1532--1543.

\bibitem[{Rashkin et~al.(2019)Rashkin, Smith, Li, and
  Boureau}]{rashkin-etal-2019-towards}
Hannah Rashkin, Eric~Michael Smith, Margaret Li, and Y-Lan Boureau. 2019.
\newblock \href {https://doi.org/10.18653/v1/P19-1534} {Towards empathetic
  open-domain conversation models: A new benchmark and dataset}.
\newblock In \emph{Proceedings of the 57th Annual Meeting of the Association
  for Computational Linguistics}, pages 5370--5381, Florence, Italy.
  Association for Computational Linguistics.

\bibitem[{Ruan and Ling(2021)}]{Ruan_2021}
Yu-Ping Ruan and Zhenhua Ling. 2021.
\newblock Emotion-regularized conditional variational autoencoder for emotional
  response generation.
\newblock \emph{{IEEE} Transactions on Affective Computing}, pages 1--1.

\bibitem[{Serban et~al.(2017)Serban, Sordoni, Lowe, Charlin, Pineau, Courville,
  and Bengio}]{Serban_Sordoni_Lowe_Charlin_Pineau_Courville_Bengio_2017}
Iulian Serban, Alessandro Sordoni, Ryan Lowe, Laurent Charlin, Joelle Pineau,
  Aaron Courville, and Yoshua Bengio. 2017.
\newblock A hierarchical latent variable encoder-decoder model for generating
  dialogues.
\newblock \emph{Proceedings of the AAAI Conference on Artificial Intelligence},
  31(1).

\bibitem[{Serban et~al.(2016)Serban, Sordoni, Bengio, Courville, and
  Pineau}]{10.5555/3016387.3016435}
Iulian~V. Serban, Alessandro Sordoni, Yoshua Bengio, Aaron Courville, and
  Joelle Pineau. 2016.
\newblock Building end-to-end dialogue systems using generative hierarchical
  neural network models.
\newblock In \emph{Proceedings of the Thirtieth AAAI Conference on Artificial
  Intelligence}, AAAI'16, page 3776–3783. AAAI Press.

\bibitem[{Shen et~al.(2021)Shen, Baevski, Morcos, Keutzer, Auli, and
  Kiela}]{shen-etal-2021-reservoir}
Sheng Shen, Alexei Baevski, Ari Morcos, Kurt Keutzer, Michael Auli, and Douwe
  Kiela. 2021.
\newblock Reservoir transformers.
\newblock In \emph{Proceedings of the 59th Annual Meeting of the Association
  for Computational Linguistics and the 11th International Joint Conference on
  Natural Language Processing (Volume 1: Long Papers)}, pages 4294--4309,
  Online. Association for Computational Linguistics.

\bibitem[{Shen et~al.(2018)Shen, Su, Niu, and
  Demberg}]{10.5555/3504035.3504704}
Xiaoyu Shen, Hui Su, Shuzi Niu, and Vera Demberg. 2018.
\newblock Improving variational encoder-decoders in dialogue generation.
\newblock In \emph{Proceedings of the Thirty-Second AAAI Conference on
  Artificial Intelligence and Thirtieth Innovative Applications of Artificial
  Intelligence Conference and Eighth AAAI Symposium on Educational Advances in
  Artificial Intelligence}, AAAI'18/IAAI'18/EAAI'18. AAAI Press.

\bibitem[{Shi et~al.(2021)Shi, Katuwal, Suganthan, and Tanveer}]{SHI2021107978}
Qiushi Shi, Rakesh Katuwal, P.N. Suganthan, and M.~Tanveer. 2021.
\newblock Random vector functional link neural network based ensemble deep
  learning.
\newblock \emph{Pattern Recognition}, 117:107978.

\bibitem[{Song et~al.(2019)Song, Zhang, Cui, Wang, and
  Liu}]{Song2019ExploitingPI}
Haoyu Song, Weinan Zhang, Yiming Cui, Dong Wang, and Ting Liu. 2019.
\newblock Exploiting persona information for diverse generation of
  conversational responses.
\newblock In \emph{IJCAI}.

\bibitem[{Sun et~al.(2021)Sun, Feng, Li, Liu, and
  Li}]{sun-etal-2021-generating}
Bin Sun, Shaoxiong Feng, Yiwei Li, Jiamou Liu, and Kan Li. 2021.
\newblock Generating relevant and coherent dialogue responses using
  self-separated conditional variational {A}uto{E}ncoders.
\newblock In \emph{Proceedings of the 59th Annual Meeting of the Association
  for Computational Linguistics and the 11th International Joint Conference on
  Natural Language Processing (Volume 1: Long Papers)}, pages 5624--5637,
  Online. Association for Computational Linguistics.

\bibitem[{Tay et~al.(2020)Tay, Bahri, Metzler, Juan, Zhao, and
  Zheng}]{https://doi.org/10.48550/arxiv.2005.00743}
Yi~Tay, Dara Bahri, Donald Metzler, Da-Cheng Juan, Zhe Zhao, and Che Zheng.
  2020.
\newblock Synthesizer: Rethinking self-attention in transformer models.

\bibitem[{Tevet and Berant(2021)}]{tevet-berant-2021-evaluating}
Guy Tevet and Jonathan Berant. 2021.
\newblock \href {https://doi.org/10.18653/v1/2021.eacl-main.25} {Evaluating the
  evaluation of diversity in natural language generation}.
\newblock In \emph{Proceedings of the 16th Conference of the European Chapter
  of the Association for Computational Linguistics: Main Volume}, pages
  326--346, Online. Association for Computational Linguistics.

\bibitem[{Tolstikhin et~al.(2017)Tolstikhin, Bousquet, Gelly, and
  Schoelkopf}]{https://doi.org/10.48550/arxiv.1711.01558}
Ilya Tolstikhin, Olivier Bousquet, Sylvain Gelly, and Bernhard Schoelkopf.
  2017.
\newblock Wasserstein auto-encoders.

\bibitem[{Ueyama and Kano(2020)}]{ueyama-kano-2020-diverse}
Ayaka Ueyama and Yoshinobu Kano. 2020.
\newblock Diverse dialogue generation with context dependent dynamic loss
  function.
\newblock In \emph{Proceedings of the 28th International Conference on
  Computational Linguistics}, pages 4123--4127, Barcelona, Spain (Online).
  International Committee on Computational Linguistics.

\bibitem[{Vaswani et~al.(2017)Vaswani, Shazeer, Parmar, Uszkoreit, Jones,
  Gomez, Kaiser, and Polosukhin}]{NIPS2017_3f5ee243}
Ashish Vaswani, Noam Shazeer, Niki Parmar, Jakob Uszkoreit, Llion Jones,
  Aidan~N Gomez, \L~ukasz Kaiser, and Illia Polosukhin. 2017.
\newblock Attention is all you need.
\newblock In \emph{Advances in Neural Information Processing Systems},
  volume~30. Curran Associates, Inc.

\bibitem[{Wang et~al.(2021)Wang, Zheng, Jiang, and
  Huang}]{wang-etal-2021-diversifying}
Yida Wang, Yinhe Zheng, Yong Jiang, and Minlie Huang. 2021.
\newblock Diversifying dialog generation via adaptive label smoothing.
\newblock In \emph{Proceedings of the 59th Annual Meeting of the Association
  for Computational Linguistics and the 11th International Joint Conference on
  Natural Language Processing (Volume 1: Long Papers)}, pages 3507--3520,
  Online. Association for Computational Linguistics.

\bibitem[{Wang et~al.(2020)Wang, Si, Lei, and Yang}]{Wang_Si_Lei_Yang_2020}
Yiru Wang, Pengda Si, Zeyang Lei, and Yujiu Yang. 2020.
\newblock Topic enhanced controllable cvae for dialogue generation (student
  abstract).
\newblock \emph{Proceedings of the AAAI Conference on Artificial Intelligence},
  34(10):13955--13956.

\bibitem[{Wiher et~al.(2022)Wiher, Meister, and
  Cotterell}]{10.1162/tacl_a_00502}
Gian Wiher, Clara Meister, and Ryan Cotterell. 2022.
\newblock \href {https://doi.org/10.1162/tacl_a_00502} {{On Decoding Strategies
  for Neural Text Generators}}.
\newblock \emph{Transactions of the Association for Computational Linguistics},
  10:997--1012.

\bibitem[{Wu et~al.(2020)Wu, Li, Wang, Chen, Wong, Feng, Huang, and
  Wang}]{wu-etal-2020-guiding}
Bowen Wu, MengYuan Li, Zongsheng Wang, Yifu Chen, Derek~F. Wong, Qihang Feng,
  Junhong Huang, and Baoxun Wang. 2020.
\newblock Guiding variational response generator to exploit persona.
\newblock In \emph{Proceedings of the 58th Annual Meeting of the Association
  for Computational Linguistics}, pages 53--65, Online. Association for
  Computational Linguistics.

\bibitem[{Xu et~al.(2020)Xu, Liu, Yang, Raffel, and
  Niethammer}]{https://doi.org/10.48550/arxiv.2007.13003}
Zhenlin Xu, Deyi Liu, Junlin Yang, Colin Raffel, and Marc Niethammer. 2020.
\newblock Robust and generalizable visual representation learning via random
  convolutions.

\bibitem[{Zhao et~al.(2017)Zhao, Zhao, and Eskenazi}]{zhao-etal-2017-learning}
Tiancheng Zhao, Ran Zhao, and Maxine Eskenazi. 2017.
\newblock Learning discourse-level diversity for neural dialog models using
  conditional variational autoencoders.
\newblock In \emph{Proceedings of the 55th Annual Meeting of the Association
  for Computational Linguistics (Volume 1: Long Papers)}, pages 654--664,
  Vancouver, Canada. Association for Computational Linguistics.

\end{thebibliography}
\appendix

\appendix
\onecolumn
\section{Appendix}
\label{sec:appendix}
\subsection{Size comparison with the RL Transformer}

The encoder and decoder in the standard transformer and our PaRaFormer consists of an attention network, 2 layer normalization layers, and a feed forward network. For a single attention head, the number of parameters in each component can be formulated as:
\begin{equation}
    num_{attn} = 4*(n*d_{qkv})
\end{equation}
\begin{equation}
    num_{norm} = 2*(n*n)
\end{equation}
\begin{equation}
    num_{ff} = (n*d_{ff} + d_{ff}) + (d_{ff}*n + n)
\end{equation}
where $num_{attn}$,$num_{norm}$, and $num_{ff}$ refer to the number of parameters in the attention, layer norm, and feed forward networks respectively. $n$ refers to the.

The RL transformer consists of encoders and decoders which utilize RL self-attention networks and RL feed forward networks. For the RL self-attention network, additional randomized layers are inserted to attain the $Q$, $K$, $V$ matrices and the final output representation. Each randomized layer precedes a trainable layer which accepts both the output of the randomized layer and the original representation as input:
\begin{equation}
    num_{rl-attn} = 3 * (n*d_{r}) + 3 * ((d_{r}+n)*d_{qkv}) + (d_{qkv}*d_{r}) + ((d_{r}+d_{qkv})*n) 
\end{equation}
where $num_{rl-attn}$ refers to the number of parameters in the RL self-attention network, $d_{r}$ represents the size of the randomized layer, and $d_{qkv}$ denotes the dimensions of $Q$, $K$, $V$. On the other hand, for the RL feed forward network, no additional randomized layers are introduced. The first linear layer is regarded as a randomized layer and the second linear layer accepts the output of the randomized layer and the original representation as input:
\begin{equation}
    num_{rl-ff} = (n*d_{ff}+d_{ff}) + ((d_{ff}+n)*n + n))
\end{equation}
where $num_{rl-ff}$ refers to the number of parameters in the RL feed forward network. In this case, the size of the randomized layer is 4 times the size of the randomized layer used in the RL self-attention network.

In the original implementation, where $n = 300$ and $d_{r} = 512$, the RL attention network comprises 1,030,144 parameters, while the attention network of the PaRaFormer consists of 153,600 parameters. Furthermore, the feed-forward network in the PaRaFormer is composed of 1,231,148 parameters, whereas the feed-forward network in the RL Transformer contains 1,321,148 parameters. Hence, it is apparent that there is a significant size disparity between the RL Transformer and the PaRaFormer, especially in the RL attention network. The smaller size of the PaRaFormer indicates that it would demand substantially fewer computational resources in comparison.

\subsection{Kaiming Weight Initialization Constraint}
For an arbitrary layer $i$ in a neural network, the layer output $y_{i}$ can be expressed as: 

\begin{equation}
    y_{i} = W_{i}^{0}X_{0} + W_{i}^{1}X_{1} + W_{i}^{2}X_{2} + \cdots + W_{i}^{N_{i}}X_{N_{i}}
\end{equation}
where $X$ and $W$ refer to the layer input and layer weights respectively, and $N_{i}$ represents the size of the input to layer $i$. Subsequently, the variance of the output $y^{i}$ can be derived via the following equation:

\begin{equation}
\begin{split}
    y^{i} = & Var(W_{i}^{0}X_{0} + W_{i}^{1}X_{1} + W_{i}^{2}X_{2} + \cdots + W_{i}^{N_{i}}X_{N_{i}}) \\ 
    = &  N_{i}*Var(\textbf{W}_{i}\textbf{X}_{i}) \\
    = & N_{i}*Var(\textbf{W}_{i})Var(\textbf{X}_{i}) + Var(\textbf{W}_{i})(E[\textbf{X}_{i}])^{2} + (E[\textbf{W}_{i}])^{2}Var(\textbf{X}_{i}])\\
    &\text{Assuming } E[\textbf{W}_{i}] = 0,\\
    = & N_{i}*Var(\textbf{W}_{i})Var(\textbf{X}_{i}) + Var(\textbf{W}_{i})(E[\textbf{X}_{i}])^{2}\\
    = & N_{i}*Var(\textbf{W}_{i})(E[\textbf{X}_{i}^{2}])
\end{split}
\end{equation}
where $\textbf{W}_{N_{i}}$ represents the weight matrix and $\textbf{X}^{N_{i}}$ represents the input vector. Then, further expanding $E[\textbf{X}_{i}^{2}]$:
\begin{equation}
\begin{split}
    E[\textbf{X}_{i}^{2}] = & \int_{-\infty}^{\infty} \textbf{X}_{i}^{2}P(\textbf{X}_{i}) d\textbf{X}_{i}\\
    = & \int_{-\infty}^{\infty} f(y_{i-1})^{2}P(y_{i-1}) dy_{i-1}\\
    &\text{For linear activation},\\
    = & \int_{-\infty}^{\infty}y_{i-1}^{2}P(y_{i-1}) dy_{i-1}\\
    = & Var(y_{i-1})
\end{split}
\end{equation}
Hence, $Var(y_{i}) = N_{i}*Var(\textbf{W}_{i}) * Var(y_{i-1})$. Then, combining all $L$ layers in the network would result in the following expression:
\begin{equation}
Var(y_{L}) = Var(y_{1})(\prod_{k=2}^{l}N_{k}Var(W_{k}))
\end{equation}
In order to prevent both the exploding and vanishing gradient problems, the variance of the input should be equivalent to the variance of the output. Hence, we arrive at the following constraint for each layer:
\begin{equation}
    N_{i}Var(W_{i}) = 1
\end{equation}
which results in the following weight initialization:
\begin{equation}
    W_{i} \sim N(0, \frac{1}{N_{i}})
\end{equation}

\subsection{Sample Responses}

\begin{table*}[!b]
\caption{Sample responses generated by transformer-based baselines.}
\centering
\scalebox{0.72}{
\begin{tabular}{l} 
\hline
\multicolumn{1}{c}{\textbf{Context}}                                                                                    \\ 
\hline
User:~How long have you been in England ?                                                        \\
Agent:~For almost two years                                       \\
User:~Almost two years ? Your English is very good .                                                                                              \\ 
\hline
\multicolumn{1}{c}{\textbf{Response}}                                                                                   \\ 
\hline
$Transformer$:~I am sorry .                                       \\
$CVAE$:~Well , there is a lot of money .                                       \\ 
$SVT$:~and ?                                       \\ 
$GPT-2$:~It's been a long time since we've had a chance to see .                                      \\ 
$RL\:Transformer$:~I am in junior high school for a year .    \\      
$PaRaFormer_{N}$:~I had a great time .                                                                         \\
$PaRaFormer_{K}$:~I was in a publishing company .                                        \\ 
\hline
\multicolumn{1}{c}{\textbf{Context}}                                                                                    \\ 
\hline
User:~I really need to start eating healthier .                                                           \\
Agent:~I have to start eating better too . \\
User:~hat kind of food do you usually eat ?      \\
Agent:~I try my best to eat only fruits ,\\ vegetables , and chicken .                 \\
User:~Is that really all that you eat ?                                       \\ 
\hline
\multicolumn{1}{c}{\textbf{Response}}                                                                                   \\
\hline
$Transformer$:~much is the same ?                                       \\
$CVAE$:~I'm hungry at the cafeteria .                                       \\ 
$SVT$:~I think you eat a lot of cooking .                                       \\ 
$GPT-2$:~It's time for a change .                                       \\ 
$RL\:Transformer$:~I don't really like cooking .   \\       
$PaRaFormer_{N}$: I really like food .            \\      
$PaRaFormer_{K}$: I think I only have a good breakfast .     \\
\hline
\end{tabular}}
\end{table*}

\end{document}